# Efficient Adaptation of LLMs for Hate Speech Detection in Low-Resource Languages: A Comparative Study on Roman Urdu

**Toneema Zubair**
*Department of Computer Sciences*
*Information Technology University*
Lahore (54000), Pakistan
toneemaz@gmail.com

**Muhammad Junaid Asif**
Artificial Intelligence Technology Centre
National Centre for Physics
Islamabad (44000), Pakistan
junaid.asif@ncp.edu.pk

**Faisal Kamiran**
*Department of Computer Sciences*
*Information Technology University*
Lahore (54000), Pakistan
faisal.kamiran@itu.edu.pk

**Hafiz Hassan Saeed**
*Department of Computer Sciences*
*Information Technology University*
Lahore (54000), Pakistan

**Rana Fayyaz Ahmad**
Artificial Intelligence Technology Centre
National Centre for Physics
Islamabad (44000), Pakistan

**Abstract – It is challenging to detect hate speech in Low Resource Languages (LRLs) because of the absence of annotated data, the informality of its language structure, and the lack of standardized grammar. A good example of such a challenge is Roman Urdu which is broadly used by South Asians on social media platforms and has a high variation while lacking contextually consistent spellings. This paper aims to conduct a comprehensive assessment of Large Language Models (LLMs) for Hate Speech Detection (HSD) in Roman Urdu script and fine-tune these LLMs using the Parameter-Efficient Fine-Tuning (PEFT) method called Low-Rank Adaptation (LoRA). To evaluate zero-shot inference, we benchmarked it against PEFT on different transformer models, including Mistral, LLaMA, Falcon, and multilingual BERT. Experiments are conducted on the PURUTT (Parallel Urdu and Roman Urdu Corpus for Toxic Comments and Transliteration) dataset with over 72,000 annotated comments. The results suggest that zero shot models perform moderately (F1 = 0.56), but updating a small fraction of the model trainable parameters improves the classification performance significantly (F1 > 0.93). Our results have shown that PEFT delivers outstanding performance alongside excellent computational efficiency, making it highly suitable for low-resource language processing tasks.**

## 1. INTRODUCTION

The quick expansion of social media platforms has transformed the way people communicate, share their perspectives, and engage in any type of general discourse. While these platforms have made information widely available, they also allow easy dissemination of harmful information including hate speech, offensive language and abusive messages [1]. Especially hate speech is very dangerous in society, as it can lead to violence, increase discrimination, and destabilize social cohesion [2]. Therefore, Hate Speech Detection (HSD) is a critical research problem, particularly within the area of Natural Language Processing (NLP).

Over the past decade, hate speech detection has made substantial advances particularly for languages with extensive resources, including English. These developments have resulted from the availability of large-annotated datasets, powerful tools for deep learning, and standardized linguistic representations [3]. In such environments, conventional machine learning approaches, sequential to deep neural networks and transformer-based architecture have achieved outstanding performance. However, these approaches are not readily available for the LRLs in which the data is sparse, linguistically diverse and not standardized. Urdu language is one example of a low-resource language (LRL) which is widely used in Pakistan and among Urdu speaking community around the world. It does not have a single required spelling or grammar, resulting in a significant amount of spelling and grammatical variations, which can make it challenging for NLP models to learn consistent word representations. The problem becomes even more challenging when hate speech manifests in a semantic manner, in which models need to delve into context to get the right meaning [4].

Detecting hate speech in Roman Urdu is challenging due to a lack of high-quality, labeled data and significant class imbalance, which may lead to the model being skewed towards non-hate content. These challenges have sparked a renewed interest in using Large Language Models (LLMs) for enhanced detection capabilities. These challenges have spurred a renewed focus on the use of Large Language Models (LLMs) for enhanced detection capabilities. They have demonstrated remarkable capabilities in capturing complex linguistic nuances and interactions, such as BERT [5], GPT [5], LLaMA [5], and Mistral-7B [5]. Typically, these are models that have been trained on large collections of data and are then fine-tuned for specific applications. The main drawback of the usual fine-tuning is its high computational expense, so we use Parameter-Efficient Fine-Tuning (PEFT) with LoRA to update low-rank matrices while keeping the original model parameters frozen [6]. This study in this regard fills the gap by comparing LoRA-based PEFT with zero-shot LLM inference for Roman Urdu Hate speech detection. The following research questions are developed to steer this study:

1. What is the performance of pre-trained Large Language Models (LLMs) in zero-shot Roman Urdu hate speech detection?
2. How effective is Parameter-Efficient Fine-Tuning (PEFT) with LoRA for improving HSD in LRLs?

The authors gratefully acknowledge the ***National Centre for Physics (NCP)*** for providing financial support toward the publication of this paper.

3. Compared to zero-shot inference, how does LoRA-based PEFT perform in terms of Accuracy, Precision, Recall and F1-score?
4. What are the pros and cons of using LoRA over full fine-tuning in terms of computational efficiency and classification accuracy?
5. Which of the following transformer-based models is the most effective and efficient for Roman Urdu hate speech detection?

The contributions of this research are as follows:

- **A comparison of six transformer-based language models** (Mistral-7B, LLaMA-308B, Falcon-7B, Gemma-2B, DeepSeek-R1 and Multilingual BERT) for RU-HSD in both zero-shot and PEFT settings.
- **An extensive study to assess the effectiveness of LoRA based Parameter-Efficient Fine-Tuning (PEFT)** for adapting a pre-trained LLM into a low-resource language and fine-tuning only a small fraction of LLM parameters, which reduces computing costs.
- **A quantitative comparison of zero-shot inference with PEFT,** showing that across all models tested, this method achieves significant improvements in terms of Accuracy, Precision, Recall and F1-score, with the best recorded F1 score of 0.9387.

The rest of the paper is organized as follows: Related work is reviewed in Section II, which introduces the data set, the methodology, and experimental setup. Section IV discusses the results. The findings and implications, conclusion and future research directions are presented in Sections V and VI.

# 2. RELATED WORK

## 2.1. Traditional Machine and Deep Learning Approaches

Hate speech detection (HSD) has been extensively tackled by traditional machine learning (ML) approaches. Nasir et al. [7] tested six classifiers on HS-RU-20 Roman Urdu dataset with the best results of Logistic Regression at 81% and 87% accuracy for neutral vs hostile and offensive vs hate speech classifications respectively. In the case of English tweets, Naïve Bayes gave up to 93.06% accuracy as reported by Alaoui et al. [8] and Random Forest with TF-IDF N-gram features reported by Haider et al. [9] achieved accuracy of 90.26%. Boishakhi et al. [10] also showed that a multimodal HSD is advantageous and outlined the drawbacks of text-based approaches. In summary, ML offers a solid basis for HSD but is constrained in the ability to process low-resource languages, to generalize across languages, and to process subtle contextual differences, which all warrant consideration of deep learning and LLM-based methods.

In recent years, deep learning approaches have been introduced to boost HSD performance by learning contextual and semantic features directly from data. Miran and Yahia [11] showed that CNN-based models are effective for detecting hate-speech in English and noted the limitations of data sparsity and cross-lingual generalization. Dwivedy and Roy [12] used LSTM with multimodal cues and transfer learning to highlight the significance of multimodal cues. For Roman Urdu Saeed et al. [13] used the PURUT dataset, with the ensemble learning method getting 86.35% F1-score. Bilal et al. [14] also added the attention mechanism, Word2Vec embeddings and lexical normalization, yielding an F1-score of 88.5% in RU-HSD-30K and robustness in cross domain data.

## 2.2. Transformer-Based Models

Transformer-based models, especially BERT, have made it much more effective to detect multilingual hate speech. In the HSD-2Lang 2024 challenge, Barkhodar et al. [15] optimized, preprocessed and balanced data for Arabic and Turkish HSD to show the effectiveness of BERT for morphologically rich language. Jahan et al. [16] discussed the augmentation techniques and suggested a BERT-based contextual cosine similarity filtering that involves less change in the labels and provided better F1-score. Alatawi et al. [17] tested and contrasted the performance of Bi-LSTM and BERT, obtaining the F1 score of 96% for BERT on balanced data. As for the Twitter data, Bayrak et al. [18] used BERT-Base model for Turkish Twitter and got 92.53% accuracy for real-time moderation. On the ETHOS dataset, BERT embeddings were found by Rajput et al. [19] to be superior to FastText and GloVe.

All these studies have shown BERT to be effective, especially for morphologically complex languages and low-resource languages. There has been limited research on Roman Urdu in the context of hate speech detection, along with only a few comparisons of zero-shot, prompt-based, and parameter-efficient fine-tuning (PEFT) methods. The compromise between computational speed and accuracy of the model needs further elucidation. We use PEFT (Parameter Efficient Fine Tuning) techniques, including LoRA, adapters, and prefix tuning, to minimize computational requirements without compromising performance by updating fewer parameters in the model. Their application in Roman Urdu, however, shows that there is a gap in research for the development of efficient and scalable hate detection systems.

# 3. MATERIALS AND METHODS

**3.1. Dataset Details:** The PURUTT (Parallel Urdu and Roman Urdu Corpus for Toxic Comments and Transliteration) **[13]** corpus is a large-scale dataset for RU-HSD. Summary of PURUTT dataset is described in Table 1.

**Table 1.** Summary of the PURUTT Dataset

| Attribute | Description |
|---|---|
| **Dataset Name** | PURUTT |
| **Total Samples** | 72,771 |
| **Toxic Samples** | 13,097 |
| **Non-Toxic Samples** | 59,674 |
| **Task Type** | Binary Classification (Toxic / Non-Toxic) |
| **Language** | Roman Urdu (Latin script) |
| **Data Source** | Social media comments |
| **Class Distribution** | Imbalanced (≈18% toxic, 82% non-toxic) |
| **Key Challenges** | Non-standard spelling, phonetic variation, semantic ambiguity |
| **Preprocessing** | Tokenization, normalization, padding/truncation |
| **Data Split** | 60% Train / 20% Validation / 20% Test |
| **Handling Imbalance** | Class weighting (cost-sensitive learning) |

### 3.2. Proposed Methodology:

This paper aims to present a complete and systematic methodology ***(as shown in Fig. 2)*** for detecting hate speech in Roman Urdu, which can be implemented using LLMs with low-resource settings and computational constraints, while the complete training and evaluation procedure is in ***Algorithm 1.***

| **Algorithm 1: LoRA-Based Parameter-Efficient Fine-Tuning** |
|---|
| ***Input:*** |
| *Training Dataset D,* |
| *Pre-trained LLMs M* |
| ***Output:*** |
| *Fine-tuned Model M** |
| *1: Split **D** into training, testing and validation parts.* |
| *2: Preprocess and tokenize the **D**.* |
| *3: Compute class weights to mitigate class imbalance.* |
| *4 : Configure LoRA adapters **(r = 16, α = 8, dropout = 0.05).*** |
| *5: Freeze the parameters of **M** and insert LoRA adapters into the transformer layers.* |
| *6: Initialize the optimizer and weighted loss function.* |
| *7: **for** each training epoch do* |
| *8: Fine-tune only the LoRA parameters on the training set.* |
| *9: Evaluate the model on the validation set.* |
| *10: Save the best-performing model.* |
| *11: **end for*** |
| *12: Evaluate the best model on the test set.* |
| *13: Compute Accuracy, Precision, Recall, and F1-score.* |
| *14: **return** the fine-tuned model **M*.*** |

The proposed methodology focuses on the potential of pre-trained models for Roman Urdu NLP, yet with minimum costly fine-tuning. The informal nature of Roman Urdu, its non-standard orthography and its high phonetic variability make it a challenging language for traditional NLP techniques. Thus, a hybrid scheme that combines zero-shot inference with parameter-efficient fine-tuning by means of Low-Rank Adaptation (LoRA) is followed. Zero-shot evaluation assesses transformers' intrinsic capacity to comprehend Roman Urdu, and LoRA helps to adapt the model by modifying only a few parameters. Last, both approaches are evaluated on their classification accuracy and computational requirements to systematically compare state-of-the-art NLP techniques for low-resource language tasks.

#### 3.2.1. Baseline Approach: Zero-Shot Inference with LLMs

The first phase of the proposed framework is establishing a baseline by using zero-shot inference, where pre-trained language models were used to classify Roman Urdu hate speech without further task-specific training. The models chosen to have been trained on large-scale data sets and are multilingual. When performing inference, each Roman Urdu comment is fed to the model and predictions are made using the pre-trained representations of the model. The tokenizer is a tokenizer specific to the model; it tokenizes the text and pads or truncates it to a constant sequence length so that it can be processed in batches. An inference is made without the computation of the gradient, resulting in computational savings, and the class with the largest predicted probability is chosen. All test samples are treated in the same way to get a baseline performance. The differences between Roman Urdu and the model's pre-training data could also introduce limitations in performance, underscoring the need for strategies such as parameter-efficient fine-tuning to optimize its effectiveness for the specific tasks.

#### 3.2.2. Parameter-Efficient Fine-Tuning using LoRA

An efficient alternative to full model fine-tuning is parameter-efficient fine-tuning (PEFT), which updates only a few parameters. Methods like adapters, prefix tuning, and notably Low-Rank Adaptation (LoRA), decrease the expenses of training, and allow for task-specific adaptation. LoRA adds trainable low-rank matrices to the transformer layers, without training the original model parameters ***(as shown in Fig. 1).*** It is especially well suited for LRLs in which there is a shortage of data and computational resources. While the field of HSD has made significant progress, the use of adaptation techniques like zero-shot learning, prompt tuning and PEFT to Roman Urdu is comparatively under-explored.

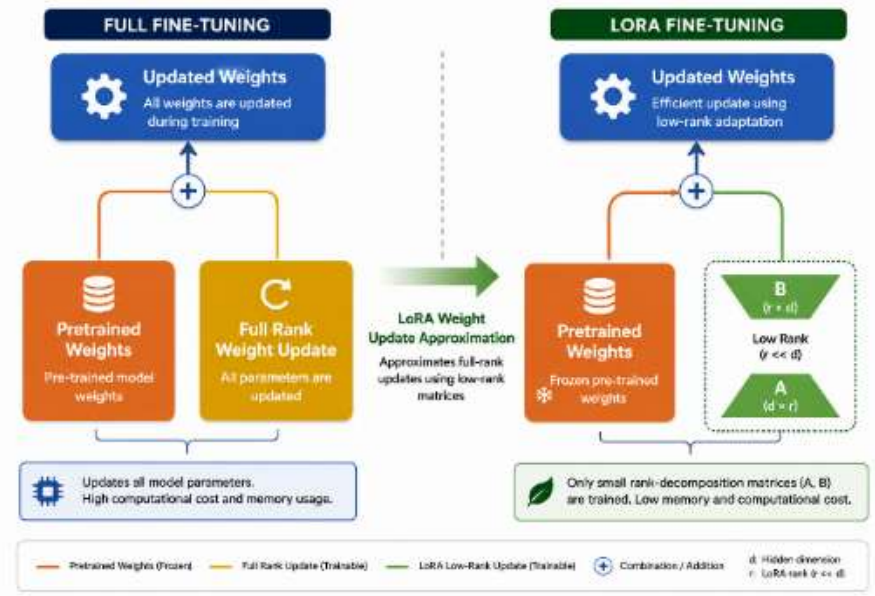

**Fig. 1.** Illustration of Complete fine-tuning versus LoRA-based PEFT, where full-rank weight updates are replaced by low-rank adaptations to efficiently update pretrained model parameters.

To investigate the trade-off between computational efficiency and classification performance, LoRA is applied to selected models for Roman Urdu hate speech detection. LoRA adds low-rank adapters to transformer projection matrices while keeping the original parameters frozen, reducing computational cost and overfitting. Mistral, LLaMA, Falcon, DeepSeek, and multilingual BERT are evaluated to compare different model sizes and architectures. Models are trained using AdamW, dropout regularization, and early stopping based on validation performance. Finally, the best-performing model is evaluated on the test set.

#### 3.2.3. Computational Efficiency

Cost-sensitive learning is applied to handle the class imbalance, with a higher penalty for the minority toxic class to ensure more attention is given to hate speech detection and to diminish the majority class's dominance. The efficiency of the computational method is realized using LoRA, which decreases the number of trainable parameters and makes it possible to fine-tune huge language models using limited resources. The proposed framework aims at comparing the zero-shot inference and PEFT based fine-tuned models to assess their predictive performance and computational efficiency. The assessment of accuracy, precision, recall and F1-score are used to evaluate models, especially F1-score because of the imbalance of classes.

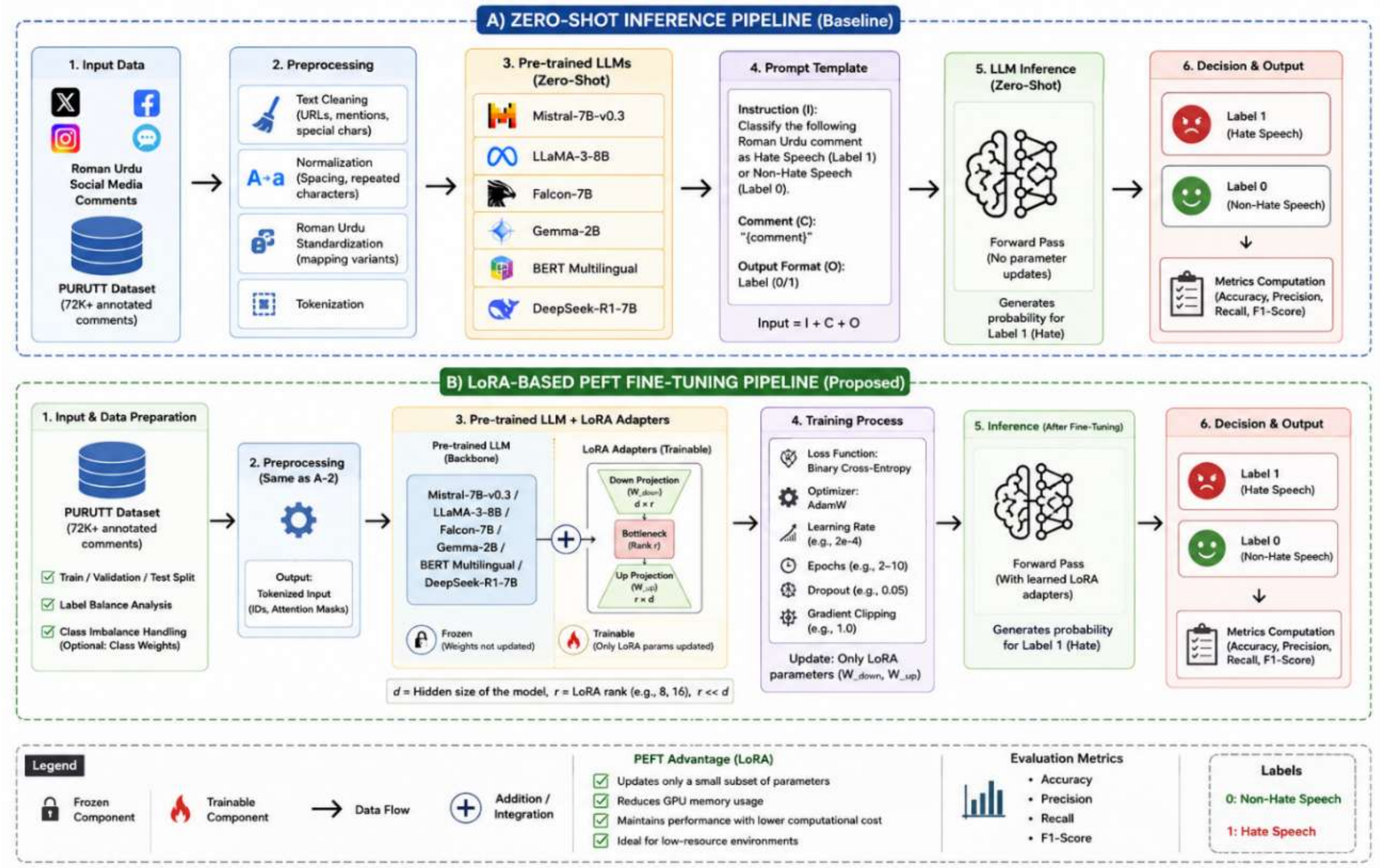

**Fig. 2.** Proposed Framework for Roman Urdu Hate Speech Detection (HSD) using Large Language Models (LLMs).

### 3.3. Implementation Details

The proposed methodology is tested and implemented with PyTorch and Hugging Face Transformers which provide scalability and reproducibility. The pre-trained models are fine-tuned with Low-Rank Adaptation (LoRA) & training and evaluation are done on NVIDIA A100 GPU.

**Table 2.** Implementation Details for LoRA-Based Fine-Tuning

| Component | Specification |
|---|---|
| **Framework** | PyTorch, Hugging Face Transformers |
| **Hardware** | NVIDIA A100 GPU |
| **Tokenization** | Model-specific tokenizers |
| **Max Sequence Length** | 512 tokens |
| **Batch Size** | 8 |
| **Optimizer** | AdamW |
| **Learning Rate** | 1 x $10^{-4}$ |
| **Epochs** | 2 |
| **Regularization** | Dropout (0.05) |
| **Fine-Tuning Method** | LoRA (PEFT) |
| **LoRA Rank (r)** | 16 |
| **LoRA Alpha** | 8 |
| **LoRA Dropout** | 0.05 |
| **Trainable Parameters** | ~7 million |
| **Evaluation Strategy** | Per epoch (validation set) |
| **Quantization** | Bit and Byte 4-bit NF4 |
| **Random seed** | 42 |

## 4. RESULTS AND DISCUSSION

The proposed method is based on two approaches: 1) zero-shot inference and 2) parameter-efficient fine-tuning (PEFT) with LoRA for Roman Urdu Hate Speech detection. Performance is assessed using various measurement indicators including accuracy, precision, and recall especially the F1- score due to the class imbalance [20] - [21].

### 4.1. Baseline Performance: Zero-Shot LLMs

The first experiment is conducted on several LLMs, without fine-tuning on the task. According to the results, zero-shot models have difficulty in learning the linguistic complexities of Roman Urdu.

**Table 3.** Performance Comparison of Zero-Shot Models on RU-HSD

| Model | Accuracy | Precision | Recall | F1-score |
|---|---|---|---|---|
| **Mistral-7B-v0.3** | 0.7262 | 0.55 | 0.56 | 0.56 |
| **DeepSeek-R1-7B** | 0.7402 | 0.51 | 0.51 | 0.51 |
| **Falcon-7B** | 0.3178 | 0.51 | 0.51 | 0.32 |
| **LLaMA-3-8B** | 0.2217 | 0.48 | 0.49 | 0.21 |
| **Gemma-2B** | 0.2221 | 0.48 | 0.49 | 0.21 |
| **BERT Multilingual** | 0.1779 | 0.46 | 0.50 | 0.15 |

As illustrated in ***Fig. 3***, the score of Mistral-7B-v0.3 is 0.56, which is the highest, while DeepSeek and Falcon-7B have scores in the range of 0.21 to 0.51. The results ***(as described in Table 3)*** indicate that pre-trained LLMs are not very effective for Roman Urdu hate speech detection without any domain adaptation. Significant differences between predictions and actual labels ***are shown in Fig. 4.***

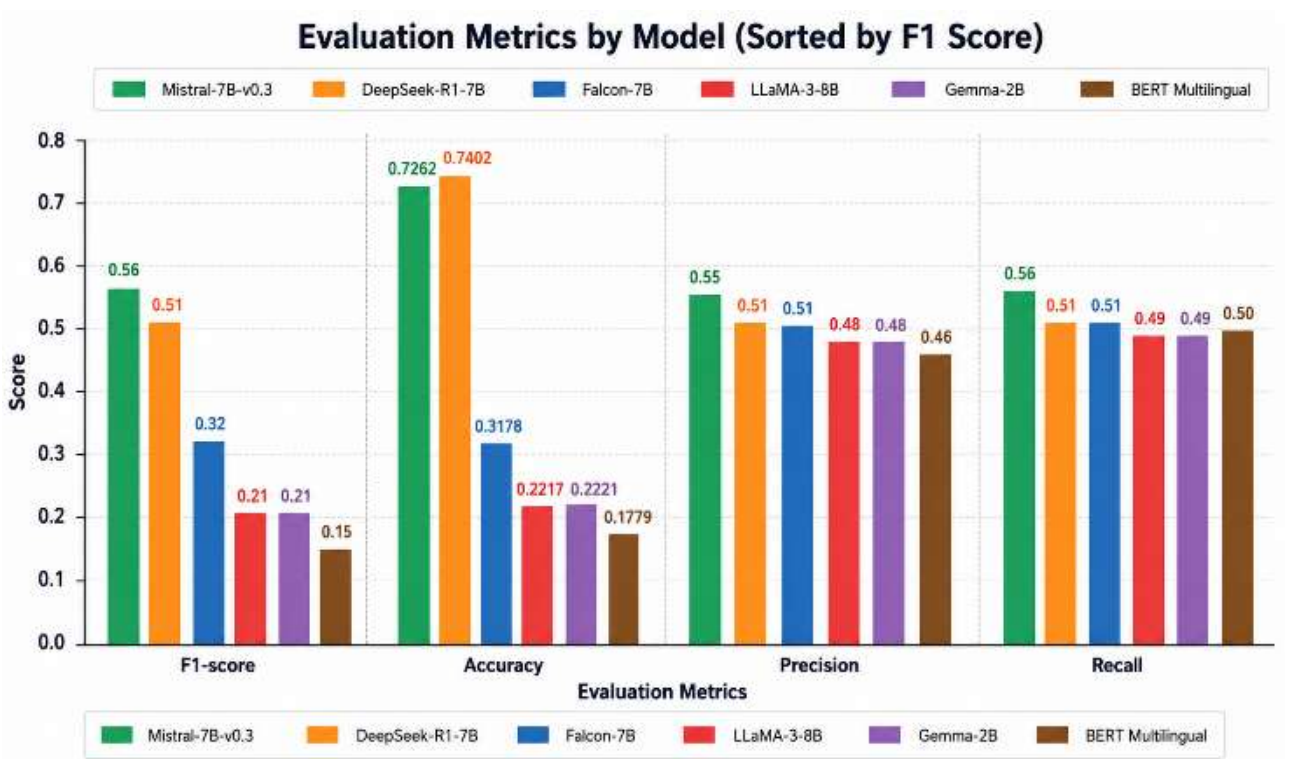

**Fig. 3.** Comparative evaluation of zero-shot models on RU-HSD across F1 Score, Accuracy, Precision, and Recall (ranked by F1 Score).

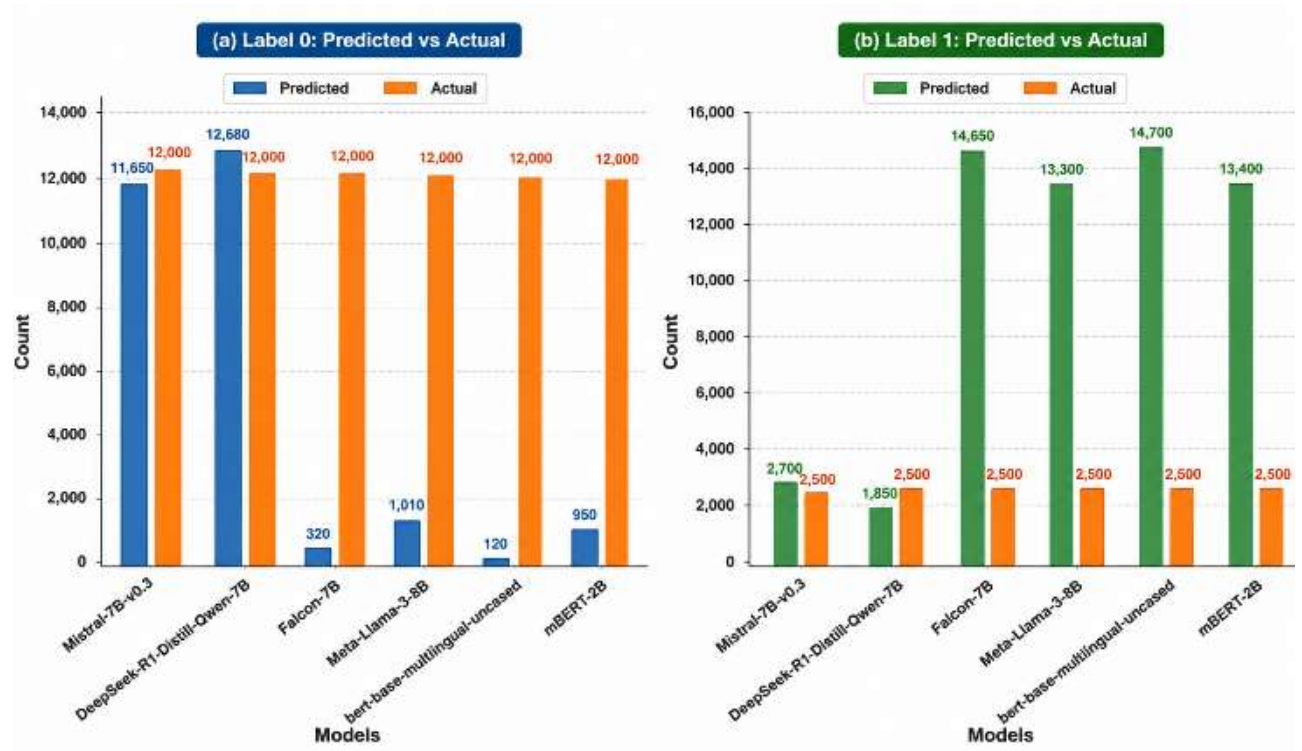

**Fig. 4.** Predicted vs. actual label distribution for hate (Label 1) and non-hate (Label 0) classes across models, highlighting class-wise prediction behavior and imbalance in zero-shot performance.

## 4.2. Performance after PEFT Fine-Tuning

The second experiment evaluates the performance boost of parameter-efficient fine-tuning (PEFT) and demonstrates a substantial improvement across all models.

**Table 4.** Performance of PEFT (LoRA) models on Roman Urdu HSD

| Model | Accuracy | Precision | Recall | F1-score |
|---|---|---|---|---|
| **Mistral-7B-v0.3** | 0.9642 | 0.9383 | 0.9392 | 0.9387 |
| **LLaMA-3-8B** | 0.9640 | 0.9407 | 0.9353 | 0.9379 |
| **Falcon-7B** | 0.8354 | 0.7278 | 0.7773 | 0.7517 |
| **Gemma-2B** | 0.9463 | 0.9052 | 0.9129 | 0.9089 |
| **BERT Multilingual** | 0.8945 | 0.8184 | 0.8222 | 0.8203 |
| **DeepSeek-R1-7B** | 0.8354 | 0.7278 | 0.7773 | 0.7517 |

***As shown in Fig. 5,*** Mistral-7B-v0.3 achieves the highest F1-score among all models, followed by LLaMA-3-8B with the score of ***0.9379***, and Gemma-2B is also above ***0.90***. The results ***(as shown in Table 4)*** show the effectiveness of PEFT to adapt the LLM to low resource tasks without changing the full LLM parameters, and how the task specific models outperform their general-purpose counterparts in most cases.

***Fig. 6*** shows that the label distributions of fine-tuned models are more aligned with the actual distribution of the labels, indicating improved calibration and reduced bias. These results are also confirmed by the confusion matrix which shows the number of false Positive and Negative values has reduced significantly, thus improving precision and recall values.

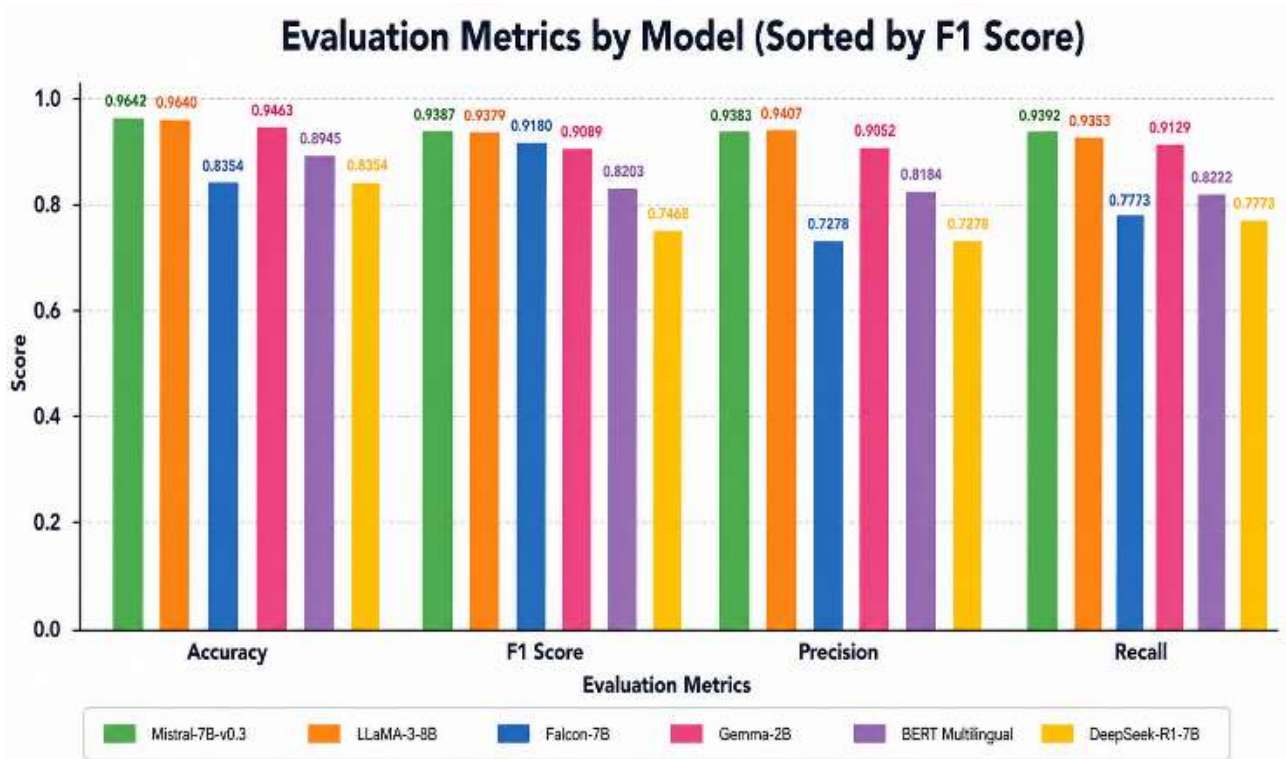

**Fig. 5.** Comparative evaluation of PEFT-enhanced models on Roman Urdu hate speech detection using Accuracy, F1 Score, Precision, and Recall (ranked by F1 Score).

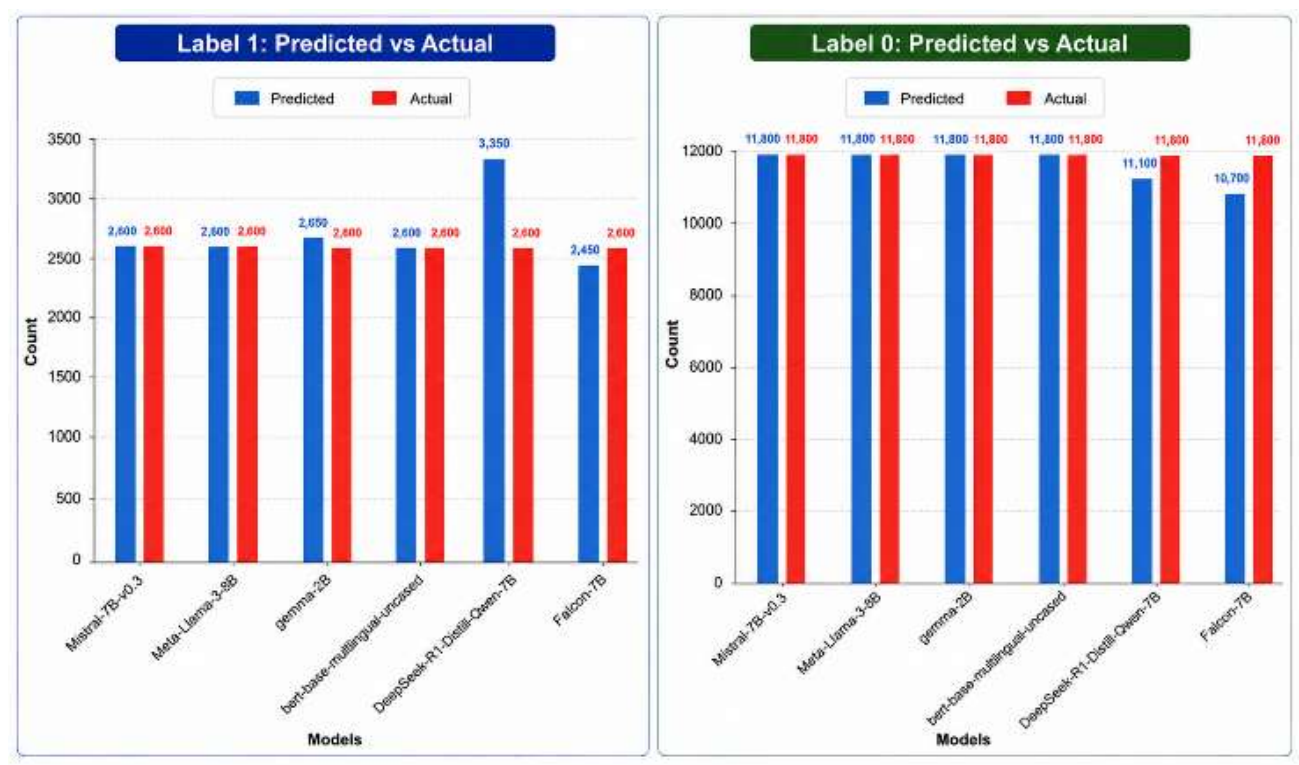

**Fig. 6.** Predicted vs. actual label distribution for hate (Label 1) and non-hate (Label 0) classes after PEFT (LoRA) fine-tuning, showing improved alignment and balanced class-wise predictions across models.

## 4.3. Comparative Analysis: Zero-Shot vs PEFT

The results ***(as presented in Table 5)*** demonstrate the efficacy of the parameter-efficient fine-tuning approach over the zero-shot result. The zero-shot is not ideal yet, ***as shown in Fig. 7***, with F1-score of approximately 0.5, whereas F1-score of approximately 0.93 after fine-tuning, indicating the necessity of adapting for the task.

**Table 5. Analysis of Zero-Shot and PEFT Performance Across Models**

| Model | Zero-Shot F1 | PEFT F1 | Improvement |
|---|---|---|---|
| **Mistral-7B** | 0.56 | 0.9387 | +0.3787 |
| **LLaMA-3** | 0.21 | 0.9379 | +0.7279 |
| **Falcon-7B** | 0.32 | 0.9180 | +0.5980 |
| **Gemma-2B** | 0.21 | 0.9089 | +0.6989 |
| **BERT Multilingual** | 0.15 | 0.8203 | +0.6703 |
| **DeepSeek-R1** | 0.51 | 0.7468 | +0.2368 |

The results show that PEFT can enhance the performance of models of various sizes and architectures, making the smaller models comparable to larger models. While PEFT is superior in terms of precision, recall, and F1-scores compared with zero-shot inference, class weighting can limit imbalance effects. Among all the models, Mistral-7B-v0.3 had the best overall performance, followed by LLaMA-3-8B, and Gemma-2B was in competitive third place. The finetuning also helped in the handling of Roman Urdu slang, abbreviations, ambiguity and code-mixing.

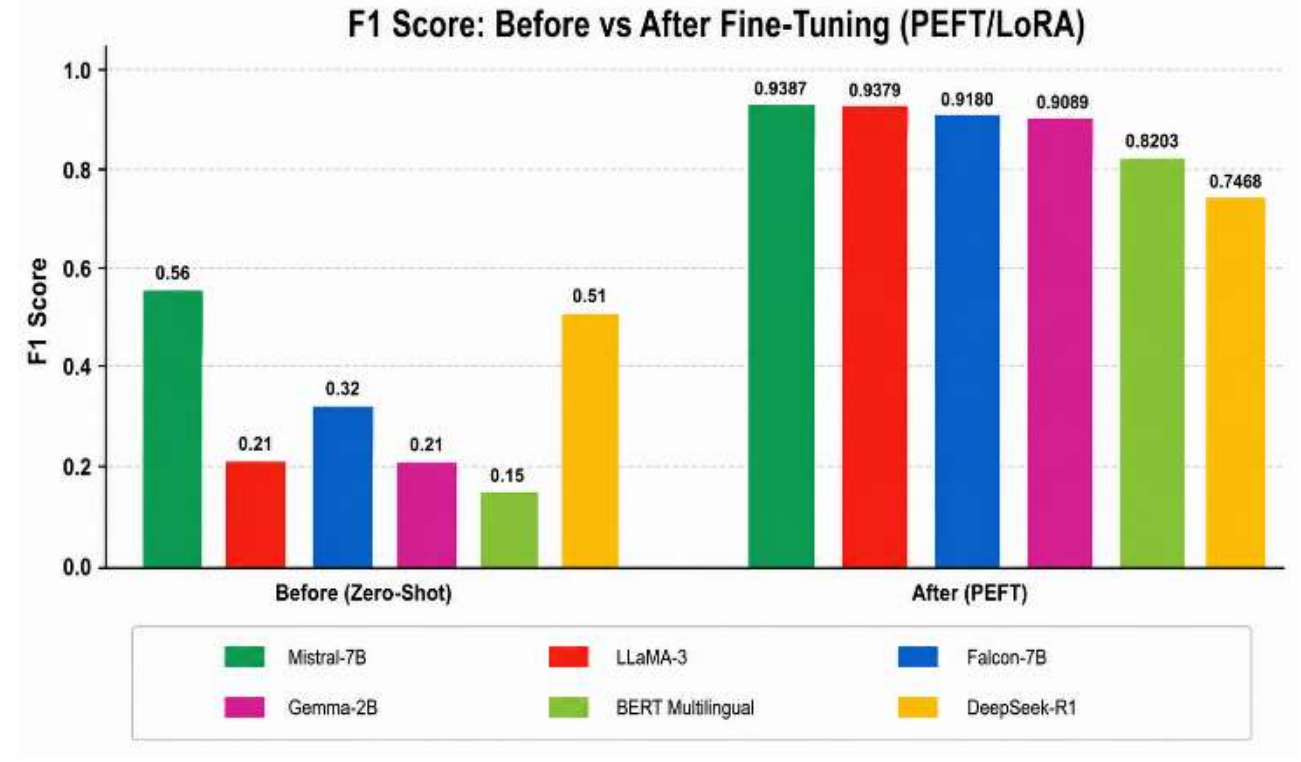

**Fig. 7.** Comparative evaluation of zero-shot and PEFT-enhanced models on Roman Urdu hate speech detection using Accuracy, F1 Score, Precision, and Recall (ranked by F1 Score).

## 5. CONCLUSIONS

The study showcases the effectiveness of Parameter-Efficient Fine-Tuning (PEFT) on Low-Rank Adaptation (LoRA) for detecting Roman Urdu hate speech, outpacing the zero-shot approach. Zero-shot LLMs show cross-lingual generalization but their accuracy is low and false positive rate on low-resource Roman Urdu is high, suggesting the need for task-specific adaptation. PEFT only updates a small subset of parameters, which means that the model can perform well without consuming a lot of resources. Models with the highest performances have F1 scores above 0.93, signifying good balance between precision and recall. Improvements are uniform for all types of transformer architectures and are generally more dramatic for larger transformers. The results show that the PEFT method is practical in the context of deployment in resource-constrained environments and significantly reduces the training costs compared to full fine-tuning. Although these are positive results, there are several limitations. The evaluation is performed on one Roman Urdu data set which might not be generalizable to other low resource languages. More nuanced types of harmful content, including different tones of harmful content, sarcasm, and implicit hate speech, are not captured by the binary classification setting. In addition, the models are mostly pre-trained in high-resource languages, which can lead to domain bias. Quantitative aspects are reported, but qualitative aspects are not discussed, such as fairness, bias, or interpretability. Lastly, cross-lingual transfer learning and domain adaptation were not considered as part of this work and would serve as interesting avenues for future research.

## References

[1] S. Kopf, "Unravelling social media critical discourse studies (SM-CDS)–four approaches to studying social media through the critical lens," *Critical Discourse Studies,* vol. 23, p. 215–232, 2026.

[2] M. A. Peters, *Limiting the capacity for hate: Hate speech, hate groups and the philosophy of hate,* vol. 54, Taylor & Francis, 2022, p. 2325–2330.

[3] A. Albladi, M. Islam, A. Das, M. Bigonah, Z. Zhang, F. Jamshidi, M. Rahgouy, N. Raychawdhary, D. Marghitu and C. Seals, "Hate speech detection using large language models: A comprehensive review," *IEEE Access,* vol. 13, p. 20871–20892, 2025.

[4] S. Tariq, T. A. Rana and F. Shahzadi, "A comparative study of sentiment analysis in Urdu and Roman Urdu: The neglected realms," *CSI Transactions on ICT,* vol. 13, p. 193–211, 2025.

[5] V. T. T. Huong and T. K. Tran, "An Assessment of Large Language," in *From Smart Cities to Smart Factories for a Sustainable Future: Proceedings of the 3rd International Conference on" From Smart Cities to Smart Factories for a Sustainable Future"(SCFF25)*, 2026.

[6] Y. Li, S. Han and S. Ji, "Vb-lora: Extreme parameter efficient fine-tuning with vector banks," *Advances in Neural Information Processing Systems,* vol. 37, p. 16724–16751, 2024.

[7] S. Nasir, A. Seerat and M. Wasim, "Hate Speech Detection in Roman Urdu using Machine Learning Techniques," in *5th International Conference on Advancements in Computational Sciences (ICACS)*, Lahore, 2024.

[8] S. S. Alaoui, Y. Farhaoui and B. Aksasse, "Hate Speech Detection Using Text Mining and Machine Learning," *International Journal of Decision Support System Technology,* vol. 14, p. 1–20, 2022.

[9] F. Haider, I. Dipty, F. Rahman, M. Assaduzzaman and A. Sohel, "Social Media Hate Speech Detection Using Machine Learning Approach," in *6th IFIP TC 12 International Conference on Computational Intelligence in Data Science (ICCIDS)*, Chennai, 2023.

[10] F. T. Boishakhi, P. C. Shill and M. G. R. Alam, "Multi-modal Hate Speech Detection using Machine Learning," *CoRR,* vol. abs/2307.11519, 2023.

[11] A. Z. Miran and H. S. Yahia, "Hate Speech Detection in Social Media (Twitter) Using Neural Network," *J. Mobile Multimedia,* vol. 19, p. 765–798, 2023.

[12] V. Dwivedy and P. K. Roy, "Deep feature fusion for hate speech detection: a transfer learning approach," *Multimedia Tools and Applications,* vol. 82, p. 36279–36301, 2023.

[13] H. H. Saeed, T. Khalil and F. Kamiran, "Urdu toxic comment classification with PURUTT corpus development," *IEEE Access,* vol. 13, p. 21635–21651, 2025.

[14] M. Bilal, A. Khan, S. Jan and S. Musa, "Context-Aware Deep Learning Model for Detection of Roman Urdu Hate Speech on Social Media Platform," *IEEE Access,* vol. 10, 2022.

[15] E. Barkhodar, I. Topçu and A. Hürriyetoglu, "Team Curie at HSD-2Lang 2024: Hate Speech Detection in Turkish and Arabic Tweets using BERT-based models," in *Proceedings of the 7th Workshop on Challenges and Applications of Automated Extraction of Socio-political Events from Text (CASE)*, 2024.

[16] M. S. Jahan, M. Oussalah, D. R. Beddiar, J. K. Mim and N. Arhab, "A Comprehensive Study on NLP Data Augmentation for Hate Speech Detection: Legacy Methods, BERT, and LLMs," *CoRR,* vol. abs/2404.00303, 2024.

[17] H. S. Alatawi, A. Alhothali and K. Moria, "Detection of Hate Speech using BERT and Hate Speech Word Embedding with Deep Model," *Applied Artificial Intelligence,* vol. 37, 2023.

[18] S. Bayrak, A. Karaca, F. Toson, A. Kocabey and F. B. Arslanoglu, "Detection of Hate Speech in Turkish Social Media Posts with BERT-Base Model," in *31st Signal Processing and Communications Applications Conference (SIU)*, 2023.

[19] G. Rajput, N. S. Punn, S. K. Sonbhadra and S. Agarwal, "Hate Speech Detection Using Static BERT Embeddings," in *Big Data Analytics - 9th International Conference, BDA 2021*, 2021.

[20] M. J. Asif, "Crowd Scene Analysis Using Deep Learning Techniques," *arXiv preprint arXiv:2505.08834,* 2025.

[21] M. J. Asif, M. S. Rafaqat, U. Nazakat, U. Khan and R. F. Ahmad, "Towards Automated Solar Panel Integrity: Hybrid Deep Feature Extraction for Advanced Surface Defect Identification," *arXiv preprint arXiv:2604.10969,* 2026.